\pdfoutput=1

\documentclass[11pt]{article}

\usepackage[preprint]{acl}
\usepackage{multirow}
\usepackage{amsmath} 
\usepackage{times}
\usepackage{color}
\usepackage{latexsym}
\usepackage{booktabs}
\usepackage{amssymb}
\usepackage{xspace}

\usepackage[T1]{fontenc}

\usepackage[utf8]{inputenc}

\usepackage{microtype}

\usepackage{inconsolata}

\usepackage{graphicx}

\newcommand{\modelname}{LLM-PEE\xspace}
\title{Towards Event Extraction with Massive Types: LLM-based Collaborative Annotation and Partitioning Extraction}

\author{
 \textbf{Wenxuan Liu\textsuperscript{1,2}},
 \textbf{Zixuan Li\textsuperscript{1,2}}$^{*}$,
 \textbf{Long Bai\textsuperscript{1,2}},
 \textbf{Yuxin Zuo\textsuperscript{1,2}}, \\
 \textbf{Daozhu Xu\textsuperscript{3}}, 
 \textbf{Xiaolong Jin\textsuperscript{1,2}}$^{*}$,
 \textbf{Jiafeng Guo\textsuperscript{1,2}},
 \textbf{Xueqi Cheng\textsuperscript{1,2}}
\\
 \textsuperscript{1}School of Computer Science and Technology, University of Chinese Academy of Sciences \\
 \textsuperscript{2}Key Laboratory of Network Data Science and Technology, \\Institute of Computing Technology, Chinese Academy of Sciences
 \\\textsuperscript{3}State Key Laboratory of Geo-Information Engineering, Xi’an, Shaanxi, China
\\
 \texttt{\{liuwenxuan2024z, lizixuan, jinxiaolong\}@ict.ac.cn}
}

\newcommand \footnoteONLYtext[1]
{
	\let \mybackup \thefootnote
	\let \thefootnote \relax
	\footnotetext{#1}
	\let \thefootnote \mybackup
	\let \mybackup \imareallyundefinedcommand
}
\begin{document}

\maketitle
\footnoteONLYtext{$^{*}$Corresponding authors.}
\begin{abstract}

Developing a general-purpose extraction system that can extract events with massive types is a long-standing target in Event Extraction (EE). In doing so, the challenge comes from two aspects: 1) The absence of an efficient and effective annotation method. 2) The absence of a powerful extraction method can handle massive types. For the first challenge, we propose a collaborative annotation method based on Large Language Models (LLMs). Through collaboration among multiple LLMs, it first refines annotations of trigger words from distant supervision and then carries out argument annotation. Next, a voting phase consolidates the annotation preferences across different LLMs. Finally, we create the EEMT dataset, the largest EE dataset to date, featuring over \textbf{200,000} samples, \textbf{3,465} event types, and \textbf{6,297} role types. For the second challenge, we propose an LLM-based Partitioning EE method called \modelname. To overcome the limited context length of LLMs, LLM-PEE first recalls candidate event types and then splits them into multiple partitions for LLMs to extract events. The results in the supervised setting show that LLM-PEE outperforms the state-of-the-art methods by \textbf{5.4}\% in event detection and \textbf{6.1}\% in argument extraction. In the zero-shot setting, LLM-PEE achieves up to \textbf{12.9}\% improvement compared to mainstream LLMs, demonstrating its strong generalization capabilities.

\end{abstract}

\section{Introduction}

\begin{figure}[t]
 \centering
 \includegraphics[width=0.7\columnwidth]{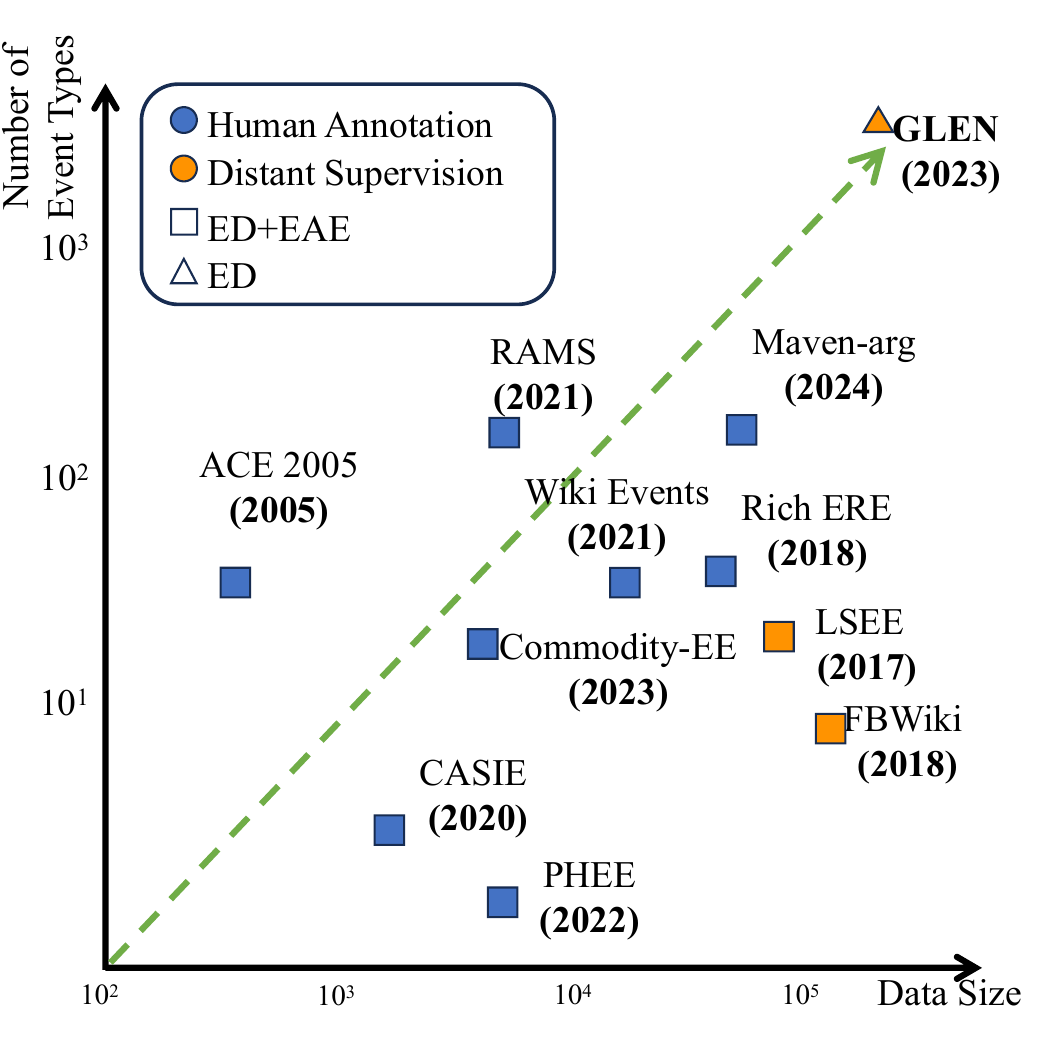}
 \caption{Statistics on the existing EE datasets.}
  \label{fig:schema}
\end{figure}

Event Extraction (EE) aims to identify structural event information from text, which contains two subtasks~\cite{Zhan_Li_Conger_Palmer_Ji_Han_2023}, i.e., Event Detection (ED) and Event Argument Extraction (EAE). The former identifies the trigger words of events (event triggers) and their corresponding types, while the latter extracts arguments and their associated roles based on a trigger and its event type. EE has demonstrated its value across a variety of domains, including finance~\cite{lee2021effective}, biomedical research~\cite{sun2022pheedatasetpharmacovigilanceevent}, and Cyber-security~\cite{satyapanich2020casie}. Each of these domains has its specific event types, which jointly form a large event schema containing massive types. It leads researchers~\cite{Bunt_Ide_Lee_Petukhova_Pustejovsky_Romary_Spaulding_Conger_Gershman_Uceda-Sosa_et} in this area to pursue a general-purpose system capable of extracting events with massive types in different domains.

In doing so, the basic challenge is the lack of an effective and efficient annotation method to construct datasets. As illustrated in Figure~\ref{fig:schema}, existing datasets can be divided into two types based on their annotation methods: human-annotated and distant supervision-based ones. Human annotation is generally effective but inefficient, requiring annotators to understand long guidelines and undergo specialized training. Consequently, the obtained datasets are often limited in terms of both event type and scale. Considering that existing semantic frame knowledge bases, e.g., FrameNet~\cite{fillmore2009frames}, Propbank~\cite{kingsbury2002treebank}, contain massive types of predicates, distant supervision-based methods~\cite{chen-etal-2017-automatically,DBLP:journals/corr/abs-1712-03665,Zhan_Li_Conger_Palmer_Ji_Han_2023} automatically annotate triggers and arguments if they have been annotated in the knowledge bases, offering a more efficient alternative. For example, GLEN~\cite{Zhan_Li_Conger_Palmer_Ji_Han_2023}, the largest ED dataset to date, uses Propbank and Wikidata~\cite{vrandevcic2014wikidata} to annotate event triggers across more than 3,000 types. However, these datasets often suffer from noise in three aspects: 1) \textbf{Unreasonable Trigger Annotation:} Some predicates in the semantic frame are not considered as events in the view of EE, leading to the incorrect annotation of irrelevant words as triggers. For example, the adverb ``voluntarily'' is annotated as a trigger in GLEN. 2) \textbf{Coarse-Grained Type Annotation:} Due to the hierarchical structure of event types, a predicate often has types with multiple granularities. cmethods struggle to assign precise fine-grained types to these predicates. For example, GLEN annotates several potential types (crime, property crime, etc) for the trigger ``crime''. 3) \textbf{Missing Argument Annotation:} Unlike triggers, where candidate triggers are relatively limited, the number of potential arguments is much larger and cannot be fully enumerated by existing knowledge bases. Thus, distant supervision-based methods often miss argument annotations when the arguments are not included in the knowledge bases.

Large Language Models (LLMs) have recently achieved significant performance improvements across many Natural Language Processing (NLP) tasks and emerge as a promising approach for annotating EE datasets~\cite{Chen_Qin_Jiang_Choi_2024}. One key challenge, however, is to mitigate the annotation bias inherent in a specific LLM. Motivated by this, we propose an LLM-based collaborative annotation method. Based on the results from distant supervision-based methods, it first performs event trigger filtering by removing irrelevant triggers. This is followed by event type refinement, which assigns more fine-grained event types for triggers based on context. Finally, it identifies the roles of arguments associated with each event and refines the original annotation with human annotation rules. After each step, multiple LLMs collaborate to generate annotations, and then a voting phase is used to unify the annotation preferences across LLMs. Finally, we obtain a new dataset, called EEMT, with over 200,000 annotated samples, covering 3,465 event types and 6,297 argument role types, which is the largest EE dataset regarding event type and scale, to the best of our knowledge.

To adapt LLMs for EE with massive types, we propose a Partitioning EE method for LLMs called \modelname, which addresses the prompt length limitation when handling massive types. LLM-PEE begins by recalling the top-k most similar event types, then divides these types into several partitions, which are assembled into prompts of LLMs. Based on these partitioning prompts, LLMs extract event triggers and their corresponding arguments. Experimental results on the EEMT dataset demonstrate that LLM-PEE outperforms the state-of-the-art models by 5.4\% in event detection and 6.1\% in argument extraction. Besides, LLM-PEE achieves up to 12.9\% F1 improvement compared to mainstream LLMs in the zero-shot setting, demonstrating its strong generalization capabilities.
 
Our contributions can be summarized as follows:

\begin{itemize}
  \item We propose an LLM-based collaborative annotation method for EE with massive types, where multiple LLMs automatically annotate events collaboratively.
   \item Based on the above method, we construct the EEMT dataset, which is the largest EE dataset to date in terms of both its coverage and scale.
  \item We propose an LLM-based Partitioning Extraction method, which significantly improves EE under supervised and zero-shot settings.
\end{itemize}

\begin{figure*}[tp]
  \centering
  \includegraphics[width=0.95\linewidth]{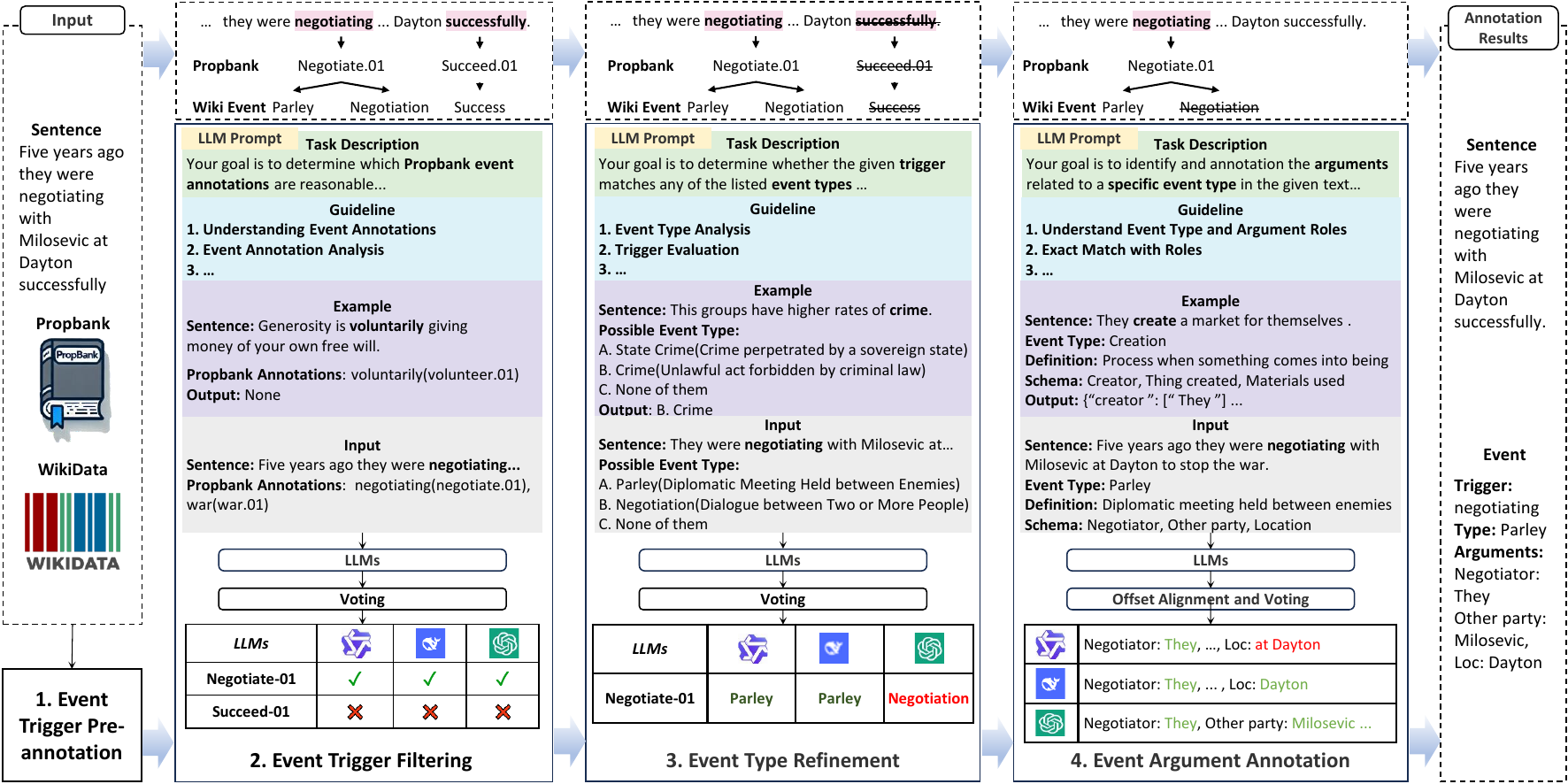}
  \caption{
  Overview of the proposed LLM-based collaborative annotation method for EE with massive types.
  }
  \label{fig:pipeline}
\end{figure*}
\section{Related Work}

\paragraph{Event Extraction Dataset.}\label{sec:event dataset}
In human annotated datasets, ACE 2005~\cite{ACE2005_DATASET} is the most commonly used dataset, including 33 event types and 22 roles.
Followed by this, Rich ERE~\cite{song-etal-2015-light} is proposed to further enhance the scale of the EE dataset.
MAVEN-arg\cite{wang2023mavenarg} is constructed based on MAVEN\cite{wang2020maven}, the current largest EAE dataset annotated by human experts with 162 event types and 612  roles. However, due to the high cost of annotating events, human annotators can not further extend schema and data scale.
For datasets from distant supervision-based methods, 
LSEE~\cite{chen-etal-2017-automatically} is constructed from FrameNet~\cite{fillmore2009frames} and WikiPedia~\cite{milne2008learning} to automatically annotate events.
GLEN~\cite{Zhan_Li_Conger_Palmer_Ji_Han_2023} is the largest ED dataset including 3465 types and 200,000 samples.
However, the dataset from distant supervision suffers from the low annotation quality.
%In addition, there are some EE datasets focused on specific domains, such as CASIE~\cite{satyapanich2020casie} for cybersecurity, PHEE~\cite{sun2022pheedatasetpharmacovigilanceevent} for the biomedical, and CommodityEE~\cite{lee2021effective} for the commodity field. The detailed statistics of the event extraction dataset are listed in Table \ref{tab:data source}.

\paragraph{Event Extraction Method.}
EE methods can be divided into two kinds: classification-based and generation-based ones. Classification-based methods~\cite{Zhang_Ji_2021, Zhan_Li_Conger_Palmer_Ji_Han_2023} tend to formulate EE as a token classification or a sequential labeling task. Generation-based methods~\cite{du2020event,liu-etal-2020-event,hsu2022degreedataefficientgenerationbasedevent} aim to generate the text containing a structured event, while these methods require manual schema-specific templates, which are difficult to be adapted to massive types. 
Nowadays, due to the strong generation abilities of LLMs, they have been widely used in EE, like InstructUIE~\cite{wang2023instructuie}, KnowCoder~\cite{li2024knowcoder}, AlignXIE~\cite{zuo2024alignxie}. Besides, some works focus on few-shot or zero-shot settings such as Code4Struct~\cite{Wang_Li_Ji_2022} and CodeIE~\cite{codeie}, which employ code format to extract information.

\section{LLM-based Collaborative Annotation}

\subsection{Annotation Method}

Given that LLMs demonstrate strong capabilities in understanding natural language, we
incorporate them into the EE annotation process. To improve efficiency, we use LLMs
to perform annotation based on trigger annotations from distant supervision,
rather than starting from scratch. By collaborating across multiple LLMs using
offset alignment and voting, we achieve consistent and precise annotation
results, making the process more effective. 

Specifically, as shown in Figure~\ref{fig:pipeline}, the proposed annotation
method consists of four steps: (1) Event trigger pre-annotation annotates the
triggers and their potential types based on distant supervised methods; (2)
Event trigger filtering filters the unreasonable event triggers via LLMs; (3)
Event type refinement maps the trigger to more fine-grained event types via
LLMs; (4) Event argument annotation further annotates the arguments via LLMs.
The detailed prompts and voting strategies are listed in Appendix~\ref{sec:prompt} and Appendix~\ref{sec:voting}. 
In the first step,
this paper takes the distant supervision-based methods based on Propbank~\cite{fillmore2009frames}
and Wikidata~\cite{vrandevcic2014wikidata}, as applied in GLEN~\cite{Zhan_Li_Conger_Palmer_Ji_Han_2023}, for example. The
proposed annotation method can be easily extended to other Distant supervision-based methods and knowledge bases.

\paragraph{Event Trigger Pre-annotation.}

Due to the massive types in the schema, directly using LLMs to annotate events
is inefficient, as it requires including extensive guidelines in the prompt of
LLMs. To improve the efficiency of the annotation process, we perform event
trigger pre-annotation using distant supervision. For example, in the distant
supervision-based method used in GLEN, a sentence is first processed by annotating
words as triggers if they are included in the Propbank predicate annotations.
Then, the event type (referred to as roleset in Propbank) of each
trigger is mapped to Wikidata QNode types via DWD
Overlay~\cite{Bunt_Ide_Lee_Petukhova_Pustejovsky_Romary_Spaulding_Conger_Gershman_Uceda-Sosa_et}.
After this step, we obtain the initial annotations of event triggers and their corresponding candidate event types.

\paragraph{Event Trigger Filtering.}

The event annotation guidelines in semantic frame knowledge bases, such as
Propbank, differ from the event definitions in mainstream EE datasets~\cite{ACE2005_DATASET}.
For example, some adverbs or adjectives are treated as event triggers in
Propbank. This gap leads to the issue of unreasonable trigger annotation, as
discussed earlier. Thus, this step leverages LLMs to filter out invalid
triggers. By observing these differences in event definition and describing them
as guidelines, we instruct LLMs to evaluate the validity of event annotations.
To help the LLMs better comprehend the task, carefully selected examples are
also included in the prompt. 

Due to potential differences and biases in the filtering process provided by
different LLMs, a voting strategy is applied to obtain the final results.
Specifically, we aggregate the valid events identified by each LLM. An event is
considered valid if the majority of LLMs support it. 
In cases where a tie occurs during the voting process, we will instruct each LLM to re-annotate the case repeatedly until the majority of LLMs support the result.

\paragraph{Event Type Refinement.}
In the original GLEN annotation, 60\% of triggers have more than one event type.
Distantly supervised methods cannot assign precise, fine-grained types to these
triggers, as this requires more accurate event definitions and a deeper semantic
understanding of the sentence. Therefore, this step uses LLMs to refine the
event types to a more fine-grained level automatically.

We formalize the event type refinement task as a multiple-choice problem for the
LLMs. Specifically, this step takes the candidate event types from Step 1 and
their corresponding descriptions as input. It then instructs the LLMs to select
the event types most accurately align with the event triggers and sentences.
Additionally, we include a ``None of them'' option for cases where none of the
candidate event types align with the event trigger. 

Each LLM performs this task independently, and the final fine-grained event type
is determined by selecting the candidate with the highest number of votes from
different LLMs. In the event of a tie during the voting process, each LLM is instructed to re-annotate the case. However, since the most probable fine-grained types typically converge within one or two candidates, such ties are generally resolved within two rounds of iterative voting.

\begin{table*}[htbp]
\footnotesize
\centering
\scalebox{0.95}{
\begin{tabular}{llrrrrrrr}

\toprule
\multicolumn{2}{l}{\textbf{Data Source}}&\textbf{Event Typess}&\textbf{Argument Types}&\textbf{Cases} &\textbf{Event Mentions}&\textbf{Argument Mentions}&\textbf{Domain}\\
\midrule 
\multicolumn{2}{l}{ACE 2005} &33&22&593&4,090&9,683&General\\
%\multicolumn{2}{l}{LSEE} &21&22&593&4,090&9,683&General\\
\multicolumn{2}{l}{CASIE} &5&26&1,594&3,027&6,135&Cybersecurity\\
%\multicolumn{2}{l}{PHEE} &2&16& 4,827&5,019&7,129&Biochem\\
\multicolumn{2}{l}{Commodity EE} &18&19&3,949&3,949&8,123&Commodity\\
\multicolumn{2}{l}{GLEN}&3,465&-&208,454&185,047&-&General\\
\multicolumn{2}{l}{Maven-arg}&162&612&4,480&98,591&290,613&General\\
\midrule 
\multicolumn{2}{l}{EEMT}&3,465&6,297&208,454&170,908&481,855&General\\

\bottomrule[1pt]
\end{tabular}}

\caption{\label{tab:data source}
Statistics of the EEMT dataset compared to those of other datasets. 
}

\end{table*}

\paragraph{Event Argument Annotation.}
After the fine-grained event type is selected, we need to annotate the event arguments based on the event type and its corresponding schemas. As mentioned above, distant supervised methods often fail to conduct argument annotations when the relevant arguments are absent from the knowledge bases. Thus, we adopt LLMs to annotate arguments and their roles in this step. However, annotating
event arguments introduces two additional challenges for LLMs: 1) \textbf{Roles
Understanding:} Event argument annotation requires LLMs to comprehensively
understand the complex event schema, including event types and role definitions.
Additionally, LLMs must be capable of analyzing syntactic structures within
sentence spans and mapping each span to a specific role. 2) \textbf{Bias in Span
Offset:} LLMs exhibit inherent bias in span offset when dealing with different
styles of text. Different LLMs are more likely to give out different offsets
in argument annotation.

Considering the additional challenges, we develop a set of guidelines for
analyzing logical relationships in sentences and assigning roles accordingly.
Then, LLMs are adopted to identify logical relationships within sentences, and
map text spans to their corresponding roles. Additionally, we include examples
of manual annotations in the prompts to help the LLM better understand the roles
of events. To eliminate bias in span offset across different LLMs, we instruct the LLMs to
refine and update the original annotation results, especially in span offset,
using the rules derived from human observations. We refer to this process as
\textbf{Offset Alignment}, which ensures greater consistency in the annotations
generated by different LLMs. Following this alignment, we employ a voting
strategy to determine the final argument annotations. For each argument in the
annotated event, if a specific argument-role pair appears in more than half of LLMs' annotation results, it is deemed valid.
If the LLMs generate completely different annotations for certain roles, we employ GPT-4o to annotate the case given the original annotation from each LLM.

\subsection{The Constructed EEMT Dataset}\label{sec:Dataset_statics}

\paragraph{Dataset Construction.} 
As some datasets already pre-annotate event triggers, we reuse the GLEN dataset,
currently the largest ED dataset with over 3,000 event types, to construct a
more comprehensive and high-quality dataset. Specifically, we employ three
mainstream, state-of-the-art LLMs for collaborative annotation:
Deepseek-V3~\cite{deepseekai2024deepseekv3technicalreport},
Qwen-Plus~\cite{qwen25}, and GPT-4o-mini~\cite{openai2024gpt4ocard}. We follow
the data splits of GLEN for the training, development, and test sets.
Additionally, we select 1,500 samples from the origin test set and create a
human-annotated test set with the help of three graduate students specializing
in NLP. More details are provided in Appendix~\ref{sec:anno detail}. 

The statistics of the EEMT dataset and the existing datasets are shown
in Table~\ref{tab:data source}. Compared to commonly used datasets, EEMT
surpasses them in terms of the size of event types, role types, and overall data
scale. Particularly, our dataset contains 10 times more event and role types
than the largest human-annotated EE dataset, MAVEN-arg. Compared to GLEN, we
filter out nearly 7.64\% of unreasonable event triggers, refine the
coarse-grained type annotation (which accounts for 61.3\% in the origin GLEN
dataset) into fine-grained ones, and annotate the arguments along with
their roles. More detailed statistics are in Appendix~\ref{sec:dataset}.

\begin{table}[h]
  \centering
  \footnotesize
  \scalebox{0.93}{
  \begin{tabular}{@{}ccccc@{}}
\toprule
 & Deepseek-V3$^{\dag}$ & Qwen-Plus$^{\dag}$ & GPT-4o-mini$^{\dag}$ & \textbf{CA} \\ \midrule
ETF & 92.1 & 91.4 & 91.6 & 93.2 \\
ETR & 95.8 & 94.9 & 94.2 & 96.2 \\
EAA & 84.7 & 83.5 & 84.2 & 85.3 \\ \bottomrule
\end{tabular}}
  \caption{Results between different single-LLMs and our collaborative annotation method(denoted as CA). ETF, ETR, EAA indicates Event Trigger Filtering, Event Type Refinement, and Event Argument Annotation, respectively. $^{\dag}$ indicate that we apply the offset alignment in EAA. We calculate the F1 score for each step. }
  \label{tab:multi}
\end{table}
\paragraph{Quality Assessment.}
To evaluate the effectiveness of the proposed annotation methods, we assess the
annotation quality of each of the three LLM-based steps on the previously
mentioned human-annotated test set. The results are in the last column of
Table~\ref{tab:multi}. The F1 scores for all steps surpass 85\%, with the event
type refinement achieving an impressive F1 score of 96.2\%. These results
strongly validate the effectiveness of the annotation method.
Additionally, we evaluate the results annotated by a single LLM. By leveraging
collaborative annotation, the F1 score improves at each step, further enhancing the quality of the annotations.

\section{LLM-based Partitioning Extraction}

\begin{figure}[t]
  \centering
  
  \includegraphics[width=0.87\columnwidth]{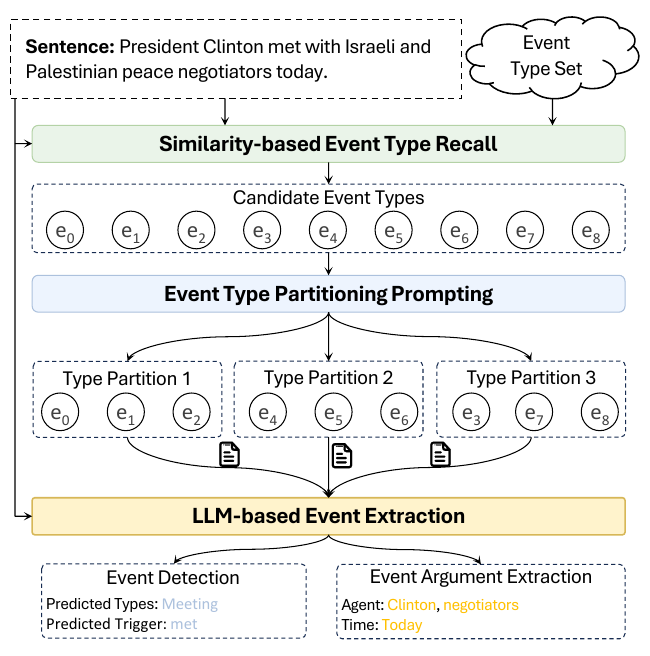}
  \caption{
  Overview of our LLM-based Partitioning Extraction Framework, including Event Detection and Event Argument Extraction.
  }
  \label{fig:model}
\end{figure}

To adopt LLMs to EE with massive types, we propose an LLM-based partitioning extraction
method for EE, called LLM-PEE. As shown in Figure~\ref{fig:model}, LLM-PEE consists
of three key components: similarity-based type recall, type-partitioning prompting,
and LLM-based event extraction. In the similarity-based type recall step, we
reduce the number of candidate event types that may appear in a sentence to a
small subset using a similarity-based model. Next, the type-partitioning prompt
divides the subset into several partitions using three different strategies, thus
further reducing the prompt length of each partition. Finally, based on the
partitioning prompts, LLMs extract event triggers and their corresponding
arguments. 

\paragraph{Similarity-based Event Type Recalling.}
Given a sentence $s={s_1,..., s_i, ..., s_n}$ and a set of candidate event types $\{e_1, ..., e_i, ..., e_m\}$, the similarity-based event type recalling step identifies the potential event types described by the sentence based on the similarity between the sentence and the event types. Following
CDEAR~\cite{Zhan_Li_Conger_Palmer_Ji_Han_2023}, we employ
ColBERT~\cite{khattab2020colbert} as the encoder of sentence and event types. Based on the embeddings from encoders, we calculate the similarities and get $k$ candidate event types for each sentence. More details are in Appendix~\ref{sec:recall}.

\paragraph{Event Type Partitioning Prompting.}
Existing LLM-based EE methods~\cite{guo2023retrieval,zhu-etal-2024-lc4ee}
typically adopt prompting learning~\cite{liu2023pre} for EE. Their input prompts generally
contain three parts, i.e., event schema information, task description, and the
input sentence. Although the number of event types is reduced to $k$
after the event type recall step, the prompt describing the event schema
information remains long, particularly for some open-source LLMs with relatively
short context lengths. Moreover, some existing
studies~\cite{liu2023lostmiddlelanguagemodels} have shown that as the prompt
length increases, the difficulty of understanding the prompt also increases.
Motivated by this, the event type partitioning prompting step splits the event types
into smaller partitions to mitigate this issue. 
To determine which event types
within a partition enhance LLM performance, we design three partitioning strategies
based on the confidence scores obtained in the similarity-based event type
recall step: 1) \textbf{Random}: The k event Types are evenly divided into N
parts randomly; 2) \textbf{Average}: The k event types are evenly divided into N
parts, ensuring that the sum of the confidences for event types in each part is
as equal as possible. This strategy aims to ensure that the extraction
difficulty is balanced across different partitions. 3) {\textbf{Level}}: Sort the k
event types based on their confidences, and then evenly divide them into N
parts. This strategy ensures different partitions have varying difficulty levels.
We analyze different partition strategies in Appendix~\ref{sec:batch}.

\begin{table*}[t]
  \centering
  \footnotesize 
  \scalebox{0.89}{
  \begin{tabular}{@{}ccccccccccccc@{}}
\toprule
\multirow{2}{*}{Method} & \multicolumn{3}{c}{TI (LLM Annotation)} & \multicolumn{3}{c}{TC (LLM Annotation)} & \multicolumn{3}{c}{TI (Human Annotation)} & \multicolumn{3}{c}{TC (Human Annotation)} \\ \cmidrule(l){2-13} 
 & P & R & F1 & P & R & F1 & P & R & F1 & P & R & F1 \\ \midrule
 
\textbf{CA}& 100.0 & 100.0 & 100.0 & 100.0 & 100.0 & 100.0 & 92.1 &	94.3 &	93.2 & 89.8 & 90.5 & 90.1 \\  \midrule
\textbf{C.B. Method} & & \multicolumn{1}{l}{} & \multicolumn{1}{l}{} &  &  &  &  &  &  &  &  &  \\
DMBERT & 50.7 & 74.2 & 60.2 & 32.2 & 47.3 & 38.3 & 49.8 & 74.0 & 59.5 & 31.7 & 47.0 & 37.9 \\
Token-Level & 55.5 & 73.3 & 63.1 & 35.5 & 49.2 & 41.2 & 56.0 & 73.1 & 63.4 & 35.9 & 48.6 & 41.3 \\
Span-Level & 53.2 & 74.7 & 62.1 & 33.5 & 50.2 & 40.5 & 53.6 & 73.5 & 62.0 & 62.0 & 49.2 & 39.5 \\
CDEAR & 60.0 & 78.2 & 67.9 & 45.1 & 55.2 & 49.7 & 61.2 & 77.7 & 68.5 & 46.3 & 54.5 & 50.1 \\ \midrule
\textbf{G.B. Method}
 &  &  &  &  &  &  &  &  &  &  &  &  \\
 
Qwen-Plus & 52.1 & 76.1 & 61.8 & 42.4 & 69.7 & 52.7 & 51.5 & 75.5 & 61.2 & 42.0 & 68.6 & 52.1 \\
GPT-4o-mini & 50.3 & 78.7 & 61.4 & 43.6 & 68.2 & 53.2 & 49.8 & 78.7 & 61.0 & 42.6 & 67.2 & 52.1 \\
Deepseek-V3 & 49.7 & \textbf{81.2} & 61.7 & 43.9 & \textbf{72.1} & 54.6 & 49.6 & \textbf{80.5} & 61.4 & 43.0 & \textbf{71.7} & 53.8 \\ 

InstructUIE & 73.1 & 75.3 & 74.2 & 53.5 & 55.5 & 54.5 & 73.5 & 74.4 & 73.9 & 53.9 & 54.9 & 54.4 \\
IEPILE & 75.7 & 71.9 & 73.8 & 56.2 & 54.2 & 55.1 & 74.8 & 72.3 & 73.5 & 55.2 & 53.5 & 54.3 \\
KnowCoder & 74.9 & 74.3 & 74.6 & 57.0 & 57.2 & 57.1 & 74.1 & 73.7 & 73.9 & 56.0 & 56.4 & 56.2 \\ \midrule

\modelname & \textbf{79.7} & 72.4 & \textbf{75.9} & \textbf{63.6} & 57.5 & \textbf{60.2} & \textbf{79.0} & 71.8 & \textbf{75.2} & \textbf{62.5} & 56.4 & \textbf{59.3} \\
\modelname w/o. e.t.r & 74.4 & 47.6 & 58.1 & 56.3 & 43.5 & 49.1 & 73.7 & 47.1 & 57.4 & 55.4 & 42.9 & 48.4 \\
\modelname w/o. t.p.p & 75.1 & 74.2 & 74.7 & 56.8 & 57.2 & 57.0 & 73.9 & 74.0 & 73.9 & 56.5 & 56.1 & 56.3 \\
\bottomrule
\end{tabular}}
  \caption{\label{tab:ED-table}
  Performance (in percentage) for the ED task on EEMT. We apply the similarity-based event type recalling (denoted as e.t.r) for all generation based models. t.p.p indicates the event type partitioning prompting.
   }
  %\label{tab:main_results}
\end{table*}

\begin{table*}[h]
  \centering
  \footnotesize  
  \scalebox{0.89}{
\begin{tabular}{@{}ccccccccccccc@{}}
\toprule
\multirow{2}{*}{Method} & \multicolumn{3}{c}{AI (LLM Annotation)} & \multicolumn{3}{c}{AC (LLM Annotation)} & \multicolumn{3}{c}{AI (Human Annotation)} & \multicolumn{3}{c}{AC (Human Annotation)} \\ \cmidrule(l){2-13} 
 & P & R & F1 & P & R & F1 & P & R & F1 & P & R & F1 \\ \midrule
 
\textbf{CA} & 100.0 & 100.0 & 100.0 & 100.0 & 100.0 & 100.0 & 90.1 & 89.5 & 89.8 & 86.4 & 84.2 & 85.3 \\  \midrule
 
\textbf{C.B. Method} & \multicolumn{1}{l}{} & \multicolumn{1}{l}{} & \multicolumn{1}{l}{} &  &  &  &  &  &  &  &  &  \\
CRF-Tagging & 25.8 & 24.9 & 25.3 & 24.4 & 23.6 & 24.0 & 24.8 & 24.7 & 24.7 & 23.8 & 23.2 & 23.5 \\
Tag-Prime & 29.1 & 25.6 & 27.3 & 27.8 & 23.4 & 25.4 & 28.8 & 25.1 & 26.8 & 27.4 & 22.8 & 24.9 \\

\midrule
\textbf{G.B. Method} 
&  &  &  &  &  &  &  &  &  &  &  &  \\ 
Qwen-Plus & 69.4 & 68.1 & 68.8 & 65.8 & 64.5 & 65.1 & 67.4 & 66.9 & 67.1 & 63.0 & 61.7 & 62.3 \\
GPT-4o-mini  & 70.4 & 67.9 & 69.1 & 66.0 & 65.0 & 65.5 & 67.9 & 66.9 & 67.4 & 63.7 & 61.2 & 62.4 \\
Deepseek-V3 & 71.9 & 68.5 & 70.2& 66.4 & 65.2 & 65.8 & 68.0 & 68.7 & 68.3 & 64.1 & 62.2 & 63.1 \\
Bart-Gen & 38.0 & 37.2 & 37.6 & 35.0 & 35.4 & 35.2 & 36.8 & 36.5 & 36.6 & 33.6 & 34.9 & 34.2 \\
InstructUIE & 65.8 & 64.7 & 65.2 & 60.1 & 59.1 & 59.6 & 64.0 & 63.6 & 63.8 & 59.1 & 57.6 & 58.3 \\
IEPILE & 66.8 & 66.5 & 66.7 & 62.1 & 59.7 & 60.9 & 66.8 & 64.2 & 65.5 & 62.1 & 59.7 & 60.9 \\
KnowCoder & 72.1 & 68.2 & 70.1 & 65.1 & 62.4 & 63.7 & 68.7 & 67.6 & 68.2 & 63.8 & 60.2 & 61.9 \\
 \midrule
\modelname & \textbf{75.5} & \textbf{70.9} & \textbf{73.1} & \textbf{69.9} & \textbf{65.6} & \textbf{67.7} & \textbf{73.2} & \textbf{68.7} & \textbf{70.9} & \textbf{67.9} & \textbf{63.8} & \textbf{65.8} \\ \bottomrule
\end{tabular} 
}
  \caption{\label{tab:EAE-table}
  Performance (in percentage) for the EAE task on EEMT.
  }
  %\label{tab:main_results}
\end{table*}

\paragraph{LLM-based EE.}
With the prompt as input, we use LLMs to conduct extraction. Specifically, we conduct
the two-stage extraction process following KnowCoder~\cite{li2024knowcoder}. For
the ED task, with the partitioning prompts as input, we generate the triggers and
corresponding event types. For the EAE task, with the golden event types and
triggers in the single sentence as input, LLMs predict potential arguments and
their corresponding roles. The details of the extraction prompt including schema, instruction and completion are
in Appendix~\ref{tab:train prompt}.

\section{Experiment}

\subsection{Experiment Setting}
\paragraph{Evaluation Metrics.}
Following GLEN~\cite{Zhan_Li_Conger_Palmer_Ji_Han_2023} and KnowCoder~\cite{li2024knowcoder}, 
we use Trigger Identification (TI) F1  and Trigger Classification (TC) F1  to evaluate ED. For EAE, we use Argument Identification (AI) F1  and Argument Classification (AC) F1 .

\paragraph{Baselines.}
For ED tasks, following GLEN, we employ four classification-based baselines(denoted as C.B. Method), including DMBERT~\cite{Wang_Han_Liu_Sun_Li_2019}, token-level classification and span-level classification, and CDEAR~\cite{Zhan_Li_Conger_Palmer_Ji_Han_2023}.
For EAE tasks, we employ two classification baselines CRF-Tagging, and Tag-Prime~\cite{hsu2023tagprime} and a generative method(denoted as G.B. Method) Bart-Gen~\cite{li2021documentlevel}. 
We compare with three LLM-based baselines, i.e., InstructUIE, IEPILE and KnowCoder. They are all fine-tuned on the proposed EEMT dataset. We also evaluate the mainstream LLMs, i.e., Qwen-Plus, GPT-4o-mini and DeepSeek-V3. For fairness, we evaluate the three models without the offset alignment in EAE. 
Besides, we conduct further evaluation on other mainstream LLMs in Appendix~\ref{sec:close-llm}.
\paragraph{Benchmark.}
For supervised evaluation, we evaluate the methods both on LLM-based testset and human annotated testset.
For zero-shot evaluation, we evaluate methods on ACE 2005~\cite{ACE2005_DATASET}.
\paragraph{Implement Details.}
We utilize LLaMA2-7B-Base~\cite{touvron2023llama2openfoundation} as the backbone and LLAMA-Factory~\cite{zheng2024llamafactory} as the training framework. Specifically, LoRA~\cite{hu2021lora} is used for efficient hyper-parameter tuning. 
We set the LoRA Rank to 8 and learning rates to 0.0003. The maximum sequence length is set to 2048, and the batch size is set to 256. Training is conducted for four epochs. We use the VLLM~\cite{kwon2023efficient} to accelerate the inference, employing greedy search and a maximum output length of 500. We recall the top 15 similar event types in the similarity-based event type recalling stage and divide them into two partitions with level strategy.

\subsection{Experiment Results}
\subsubsection{Supervised Evaluation}
The results of the supervised evaluation are listed in Tables~\ref{tab:ED-table} and ~\ref{tab:EAE-table}. We conduct a comprehensive analysis from the next four perspectives.
\paragraph{Analyses on the proposed Annotation Method.}
The collaborative annotation method (denoted as CA) is presented in the first row of Tables~\ref{tab:ED-table} and ~\ref{tab:EAE-table}.
First, the collaborative annotation method achieves an F1 score of 85\% even on the human-annotated test set, demonstrating the high effectiveness and superiority of our proposed annotation method.
Furthermore, our collaborative annotation method outperforms the direct extraction performance of any single annotation model, thereby validating the rationality and efficacy of the design underlying our method.
Specifically, in the ED task, there is a significant performance gap (90.1 v.s.56.2) between the collaborative annotation method and individual annotation models. This disparity arises because distant supervision is incorporated as auxiliary information during the annotation process, whereas directly performing event detection using individual LLMs remains highly challenging.
In the EAE task,  the performance of a single LLM significantly declines due to the absence of a collaborative mechanism and offset alignment.
This observation aligns with the results reported in Table~\ref{tab:multi}, further reinforcing the advantages of our method that mitigates bias and enhances the consistency of dataset styles.

\paragraph{Analyses on ED.}
Table~\ref{tab:ED-table} presents the results of ED on the proposed EEMT dataset. 
Compared with the KnowCoder (generation method based on LLM), LLM-PEE outperforms 1.7\% on TI and 5.4\% on TC, respectively. These results demonstrate the effectiveness of our partitioned extraction framework in addressing the event detection with massive types.
In comparison with CDEAR (designed for ED with massive types),  LLM-PEE integrates prompting learning into the framework, effectively unleashing the potential of LLMs in event detection with massive types.
Besides, without fine-tuning on the dataset, mainstream LLMs tend to over-generate events, which leads to high recall but low precision. 
LLM-PEE demonstrates superior event judgment accuracy, leading to a higher F1 score compared to the mainstream LLMs.
Additionally, the relatively low performance on TC highlights the inherent difficulty of accurately identifying event types from a large-scale event schema, which remains a significant challenge for our dataset.

\paragraph{Analyses on EAE.}

Table~\ref{tab:EAE-table} presented the results of EAE on the proposed EEMT dataset. Compared to KnowCoder, LLM-PEE achieves improvements of 3.9\% in AI and 6.2\% in AC, respectively. In \modelname, the schema representations of ED and EAE are consistent.
We guess that this consistency enables the ED task to facilitate the EAE task,
thereby deepening the model's understanding of event types and roles and contributing to higher performance on both tasks.
In comparison with the mainstream LLMs, after fine-tuning on the dataset, the argument extraction performance of LLM-PEE (trained on LLaMA2-7B) surpasses that of the original annotation LLMs. 
This improvement can be attributed to the high-quality dataset after the offset alignment and collaborative annotation.
Additionally, classification-based methods often face challenges in accurately identifying the boundaries of arguments, particularly for continuous spans (e.g., multi-token arguments) in roles such as ``purpose'', which result in lower performance in both AI and AC. That suggests generation-based methods might be more suitable for the complex EAE task.

\paragraph{Ablation Analysis.}
We conduct ablation experiments on two key modules: similarity-based event type recalling (denoted as \modelname w/o e.t.r) and event type partitioning prompting (denoted as \modelname w/o t.p.p). The results are presented in the bottom two rows of Table~\ref{tab:ED-table}.
Since the EAE task paradigm needs to specify event types and triggers in advance, we focus our ablation analysis solely on the ED task.
\modelname w/o e.t.r exhibits a significant drop of 18.3\% in TC. This result suggests that, without recalling the most similar event types, the model struggles to accurately distinguish the correct event type from the event set, particularly for unseen event types.
Furthermore, \modelname w/o t.p.p shows a performance decline of 5.1\% in TC, further validating the effectiveness of the partitioning strategies in our framework. More detailed results of the partitioning strategies are provided in Appendix\ref{sec:batch}.
Additionally, to verify the effectiveness of LLM-PEE when applied to other EE datasets, we conduct an additional supervised experiment on ACE2005. The results and analyses are presented in Appendix~\ref{sec:extra}.

\subsubsection{Zero-Shot Evaluation}
\begin{table}[h]
  \centering
  \footnotesize
  \begin{tabular}{@{}ccccc@{}}
\toprule
Method & TI & TC & AI & AC \\ \midrule
Qwen-Plus & 20.65 & 14.69 & 35.11 & 25.37 \\
GPT-4o-mini & 19.97 & 13.67 & 35.11 & 25.37 \\
Deepseek-V3 & 20.49 & \textbf{15.57} & 34.92 & 26.96 \\ \midrule
\modelname & \textbf{22.32} & 13.19 & \textbf{38.79} & \textbf{30.44} \\ \bottomrule
\end{tabular}
  \caption{Performance (in percentage) for the ED and EAE tasks on ACE 2005.}
  \label{tab:zero-shot}
\end{table}

To assess the generalization capabilities of \modelname to the unseen dataset, we conduct zero-shot experiments on the ACE 2005, which is the most commonly used EE dataset.
The results are presented in Table~\ref{tab:zero-shot}.
LLM-PEE outperforms other LLMs in TI, AI, and AC.
Notably, in TI and AI, \modelname demonstrates superior consistency in identifying spans. 
This improvement can be attributed to the high quality of the EEMT dataset, which significantly mitigates biases in span offsets and enhances the model's ability to generalize across diverse event types.
Although \modelname exhibits slightly lower performance in TC compared to other methods, 
this is because our model predicts more fine-grained event types, which are not defined in ACE 2005.
These findings convincingly demonstrate the effectiveness and advantages of \modelname, underscoring its generalization capabilities in handling complex event extraction tasks.

\section{Conclusion}
In this paper, we proposed a new LLM-based collaborative annotation method. It refines trigger annotations from distant supervision and then performs argument annotation through collaboration among multiple LLMs. We then created the new EEMT dataset based on this annotation method, which is the largest EE dataset in terms of event types and data scale. To further adapt LLMs for EE with massive types, we introduce a Partitioning EE method for LLMs called \modelname. The experimental results in both supervised and zero-shot settings demonstrate that LLM-PEE outperforms other baselines in ED and EAE tasks and even surpasses mainstream LLMs in terms of generalization capabilities in the zero-shot setting.

\section*{Limitations}
We summarized the limitations of this work and looked at them as areas for future improvement.
\begin{itemize}
    \item \textbf{Hierarchical level of event extraction}. We believe that an important factor restricting our model is that the event hierarchy is not sufficiently distinguished. We will explore how to improve the understanding of the hierarchical level of events by better event definition or positive and negative sample strategy.
    \item \textbf{End-to-End event extraction}. The proposed LLM-PEE method still divides the event extraction into two sub-tasks ED and EAE, we try to proceed directly with end-to-end event extraction based on LLMs.
    \item \textbf{Document level event extraction}. The proposed EEMT dataset is only annotated on sentence-level text, lacking document-level annotation. We hope to annotate large-scale events in document-level documents to further verify the model's capability.
\end{itemize}

\bibliography{custom}
\appendix

% \section*{Appendix}

\label{sec:appendix}

\section{LLM-based Partitioning Extraction}
\subsection{Similarity-based Event Type Recalling}\label{sec:recall}
ColBERT consists of a
BERT~\cite{devlin2018bert} layer, a convolution layer and an L2 normalization layer. In
this paper, we use $\operatorname{ColBERT}(\cdot)$ to denote the encoder.
Specifically, the embedding list of all tokens in a sentence $s$, $h_s =
[\mathbf{h}_1^s, \mathbf{h}_2^s, ...]$,  is calculated as follows:
\begin{align}
\small 
  h_s = \operatorname{ColBERT}(\text{\small{[CLS]}} \text{\small{[SENT]}} s_1, s_2, ... \text{\small{[SEP]}}),
 \end{align}
 where [SENT] is a special token indicating the object being encoded is a
 sentence.

 For an event type $e$, the corresponding embedding list of all tokens in the
 event type name, $h_e =  [\mathbf{h}_1^{e}, \mathbf{h}_2^{e}, ...]$, is
 calculated as follows:
 \begin{align}
 \small 
   h_e = \operatorname{ColBERT}(\text{\small{[CLS]}} \text{\small{[EVENT]}} \tau_i \text{\small{[SEP]}}),
 \end{align}
 where [EVENT] is a special token indicating the object being encoded is an event
 type. 

 Then, the similarity score between sentence $s$ and event type $e$ is computed
 as the sum of the maximum similarity between the token embeddings in sentence
 and event type: $\rho_{(s, e)} = \sum_{h_s}\max_{h_e}(\mathbf{h}_{i}^e, \mathbf{h}_{j}^s)$.

 A similar margin loss to CDEAR is adopted for training, which ensures that the
 best candidate is scored higher than all negative samples.

 \begin{equation}
 \small 
     \mathcal{L} = \frac{1}{N}\sum_s \sum_{e^-} \max \{ 0, (\tau - \max_{e \in C_y} \rho_{(e, s)}+  \rho_{(e^-, s)}) \}.
 \end{equation}

 Base on the similarity score, we can get $k$ candidate event types for each
 sentence.
\section{Experiment}

\subsection{Influence of partitioning strategy}\label{sec:batch}

We conduct the experiment on ED to figure out the influence of partitioning strategy in Table~\ref{tab:Batch strategy}. 
Our strategy outperforms than IEPLIE\footnote{IEPILE method builds a hard negative dictionary, however, this method leads to train-test inconsistency in our dataset. We built the hard negative dictionary and employed hard negative sampling during the training phase, and during testing, the ranked samples were randomly allocated to simulate the effect of the IEPLIE method as closely as possible.}, 
because we incorporate the information of the sentence when recalling the similar event types and ensures consistency between the training and testing phases. 
Besides, it demonstrates that the Level strategy significantly enhances the precision of event extraction. We hypothesize that sorting samples in descending order of confidence enables the grouping of the most challenging-to-distinguish event types into the same partition. This approach facilitates the model's ability to learn fine-grained distinctions between different event types, including their corresponding triggers and classifications. Notably, the inclusion of naturally occurring challenging negative samples within these partitions allows the model to better grasp the nuanced boundaries between similar or ambiguous event categories.

\begin{table}[h]
  \centering
  \small
  \begin{tabular}{lcc}
    \toprule
    \textbf{Partitioning strategy} & \textbf{TI} & \textbf{TC}\\
    \midrule
    IEPILE     & 74.9 &   59.0      \\
    Random    & 75.0 &   58.9      \\
    Average     & 75.1  &   59.1    \\
    \textbf{Level}      & \textbf{75.7}   &   \textbf{60.0}    \\
    \bottomrule
  \end{tabular}
  \caption{Results with different partitioning strategy. }
  \label{tab:Batch strategy}
\end{table}

\begin{table}[h]
\centering
\small
\begin{tabular}{@{}ccccc@{}}
\toprule
Model & TI & TC & AI & AC \\ \midrule
LLama-3.1-405B & 59.6 & 51.9 & 65.1 & 60.2 \\
Gemini-2.0-flash & 60.5 & 53.1 & 67.4 & 62.9 \\
Qwen-Turbo & 58.3 & 49.7 & 64.9 & 59.3 \\
Qwen-Plus & 61.3 & 52.8 & 67.0 & 62.3 \\
Qwen-Max & 61.3 & 53.4 & 67.5 & 62.4 \\
GPT-4o-mini & 60.1 & 52.1 & 67.4 & 62.5 \\
GPT-4o & 60.6 & 52.7 & 67.7 & 62.7 \\
Deepseek-V3 & 61.4 & 53.8 & 68.3 & 63.1 \\ \bottomrule
\end{tabular}
\caption{Results evaluated on different mainstream LLMs}
\label{tab:closed}
\end{table}
\subsection{Further Evaluation on Mainstream LLMs}\label{sec:close-llm}
To further evaluate how the mainstream LLMs perform on our dataset, we conduct the extra evaluation on our human annotation benchmark. We use the LLAMA3.1-405B-instruct~\cite{grattafiori2024llama3herdmodels}, Gemini-2.0-flash~\cite{gemini2flash}, GPT-4o, Qwen-Max to conduct further evaluation. The results are listed in ~\ref{tab:closed}.

The results suggests that without training, the mainstream LLMs exhibit almost the same annotation and extraction capabilities, which is a still challenging task for mainstream LLMs. Considering the effectiveness and efficiency, we employ the Qwen-Plus, GPT-4o-mini and DeepSeek-V3 as the final annotation LLMs.

\begin{table*}[!t]
    \centering
    \small 
    \scalebox{0.95}{
    \begin{tabular}{c|c}
    \toprule 
      Context  &  Predictions\\
    \midrule 
   \multirow{2}{40em}{ In the most recent Third \textbf{Assessment} Report ( 2001 ), IPCC wrote there is new and stronger evidence that most of the warming observed over the last 50 years is attributable to human activities. }  
   & GLEN: Assessment \\
   & EEMT: Risk Assessment \\
   \midrule 
    \multirow{2}{40em}{ The \textbf{occupying} armies existing in german territory will end soon. }  
   & GLEN: Occupation \\
   & EEMT: Military Occupation \\
   \bottomrule 
   \end{tabular}}
   \caption{Comparison of the prediction results across different datasets.}
    \label{tab:case_study}
\end{table*}
\subsection{Case Study}\label{sec:case_study}
Since we use the GLEN dataset as the original propbank annotation, we further 
analyze whether more fine-grained event type can be predicted based on the EEMT, 
we randomly selected a subset of examples from the test set for verification. 
As shown in Table \ref{tab:case_study}, we observed that compared to the original GLEN dataset, when trained on the EEMT, the model predictions shifted from ``Assessment'' to ``Risk Assessment'' and from  ``Occupation'' to    ``Military occupation'', which better aligns the sentence.
Upon analyzing the training data, we find that the key is the increase in the number of corresponding events in train set, the number of ``Risk assesement'' increases from 0 to 5, and the number of ``Military occupation'' increases from 1 to 4, which results in more accurate fine-grained event predictions.

\subsection{Supervised Evaluation on other dataset}\label{sec:extra}

\begin{table}[h]
\centering
\small
\begin{tabular}{@{}ccccc@{}}
\toprule
Model & TI & TC & AI & AC \\ \midrule
InstructUIE & 73.1 & 72.3 & 70.9 & 68.6 \\
IEPILE  & 72.9 & 72.4 & 69.9 & 68.2\\
KnowCoder & 73.1 & 72.3 & 70.9 & 68.6 \\
\modelname & 73.3 & 72.6 & 71.3& 69.2 \\
 \bottomrule
\end{tabular}
\caption{Results under the supervised evaluation in ACE 2005}
\label{tab:ace}
\end{table}
Considering the generality of LLM-PEE ~ on other datasets, we apply \modelname on the supervised setting in the dataset ACE 2005 in Table~\ref{tab:ace}. 

Compared to other fine-tuned methods, \modelname achieves state-of-the-art performance. 
However, the margin of advantage is narrower than when applied to the EEMT dataset, which contains a significantly larger number of event types. This reduction in relative performance can be attributed to the core design philosophy of our model, which primarily focuses on addressing the challenges posed by long prompts resulting from massive event types. 
When applied to datasets with fewer event types, such as ACE 2005, the inherent advantages of \modelname are somewhat diminished, as the problem it was specifically designed to tackle becomes less pronounced.

\section{Annotation Details}\label{sec:anno detail}
\subsection{Annotation Setting}
We use the DeepSeek-V3~\cite{deepseekai2024deepseekv3technicalreport}, Qwen-Plus~\cite{qwen25} and GPT-4o-mini as the annotation LLM. Considering the accuracy of the annotation and the divergence of the answers, we set the temperature to 0.5. Besides, we set a answer detection mechanism, if the obtained result can not be correctly parsed, the model is required to annotate the events agian.

\subsection{Prompts Examples}\label{sec:prompt}
We introduce the prompts we used in the Collaborative
Annotation method including Event Triggers Filtering, Event Type Refinement and Event Argument Annotation(two-stage) in Table~\ref{tab:trigger-filtering}, \ref{tab:type-refinement},~\ref{tab:argument-annotation} and ~\ref{tab:argument-preference}.

\subsection{Voting Strategies Details}\label{sec:voting}

After different stages, we employ different voting strategies. We will introduce the voting details here.

For voting after Event Trigger Filtering, if there are more than two LLMs that consider the event to be reasonable, it will consider the event to be reasonable.

For voting after Event Type Refinement, we tally and vote on the most potential event type identified by each LLM, selecting the event with the highest number of votes as the corresponding fine-level event.
In cases where a tie occurs, we apply a repeated voting procedure, capping the maximum number of voting rounds at five to prevent excessive iterations. However, in practice, ambiguous potential events were typically limited to two or three, with most reaching a definitive outcome within two voting rounds.

For voting after Event Argument Annotation, if an argument with its role appears more than or equal to half the annotation results from different LLMs, the argument is considered correct. 
Specifically, after voting, for the roles if LLMs provide three different annotation results, we instruct GPT-4o with three annotation results as a reference to generate the final annotation result.
The offset alignment prompt with multiple input are listed in Table~\ref{tab:argument-annotation-multi}.

\begin{table}[tbp]
\small
\centering
\scalebox{0.85}{
\begin{tabular}{lrrrrr}

\toprule
\multicolumn{1}{l}{\textbf{Data Source}}&\textbf{Cases} &\textbf{Event Mentions}&\textbf{Argument Mentions}\\
\midrule 
\multicolumn{2}{l}{Train} 187,468&157,051 &443,459\\
\multicolumn{2}{l}{Dev} 10,359&7,465&20,379\\
\multicolumn{2}{l}{Test}10,627&6,392&18,017\\
\multicolumn{2}{l}{Human Annotation}1,500&973&2,546\\
\bottomrule[1pt]
\end{tabular}

}

\end{table}
\section{Dataset Statistics}\label{sec:dataset}
\subsection{The split of dataset}
We introduce the spilt of our dataset in this section. For the train/dev/test set, we follow GLEN’s 90/5/5 setting, besides, to further improve the quality of our dataset. We manually annotate 1,500 cases as the human annotation set.

\subsection{Arguments Distribution}
\begin{figure}[h]
  \includegraphics[width=0.75\columnwidth]{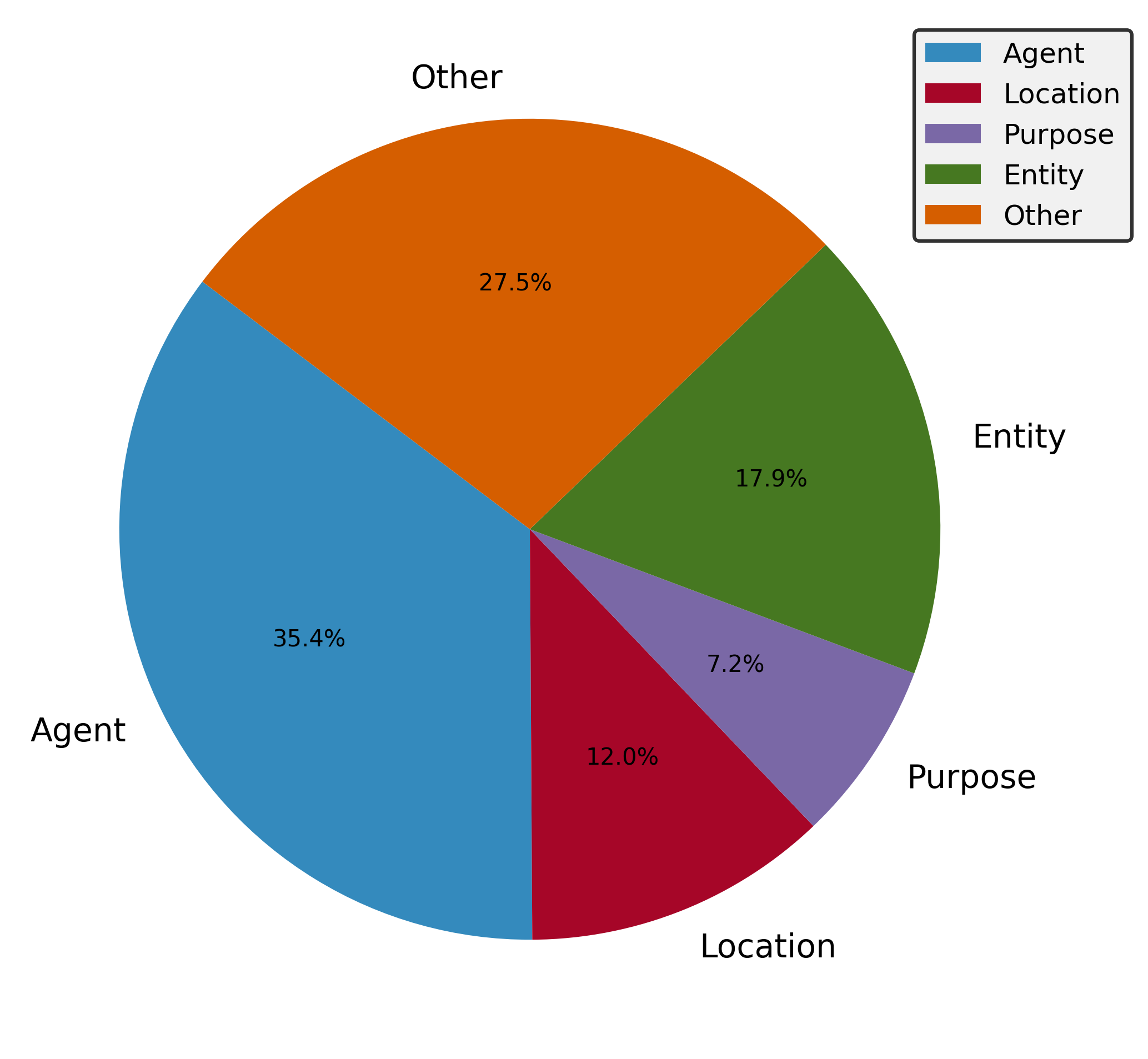}
  \caption{The argument distribution in our dataset.}
  \label{fig:argument}
\end{figure}
Furthermore, we analyze the distribution of argument in Figure~\ref{fig:argument} and categorize the role types into five broad classes: ``agent'', ``entity'', ``location'', ``purpose'', and ``others''. 
Agent, location, and entity are the three most frequently occurring argument roles, aligning with real-world distributions. 
This alignment facilitates knowledge sharing of similar roles across different event schemas, thereby reinforcing the validity of our dataset construction.

\section{Prompt for Training }\label{tab:train prompt}
\begin{figure*}[t]
   \centering
   \includegraphics[width=.9999\linewidth]{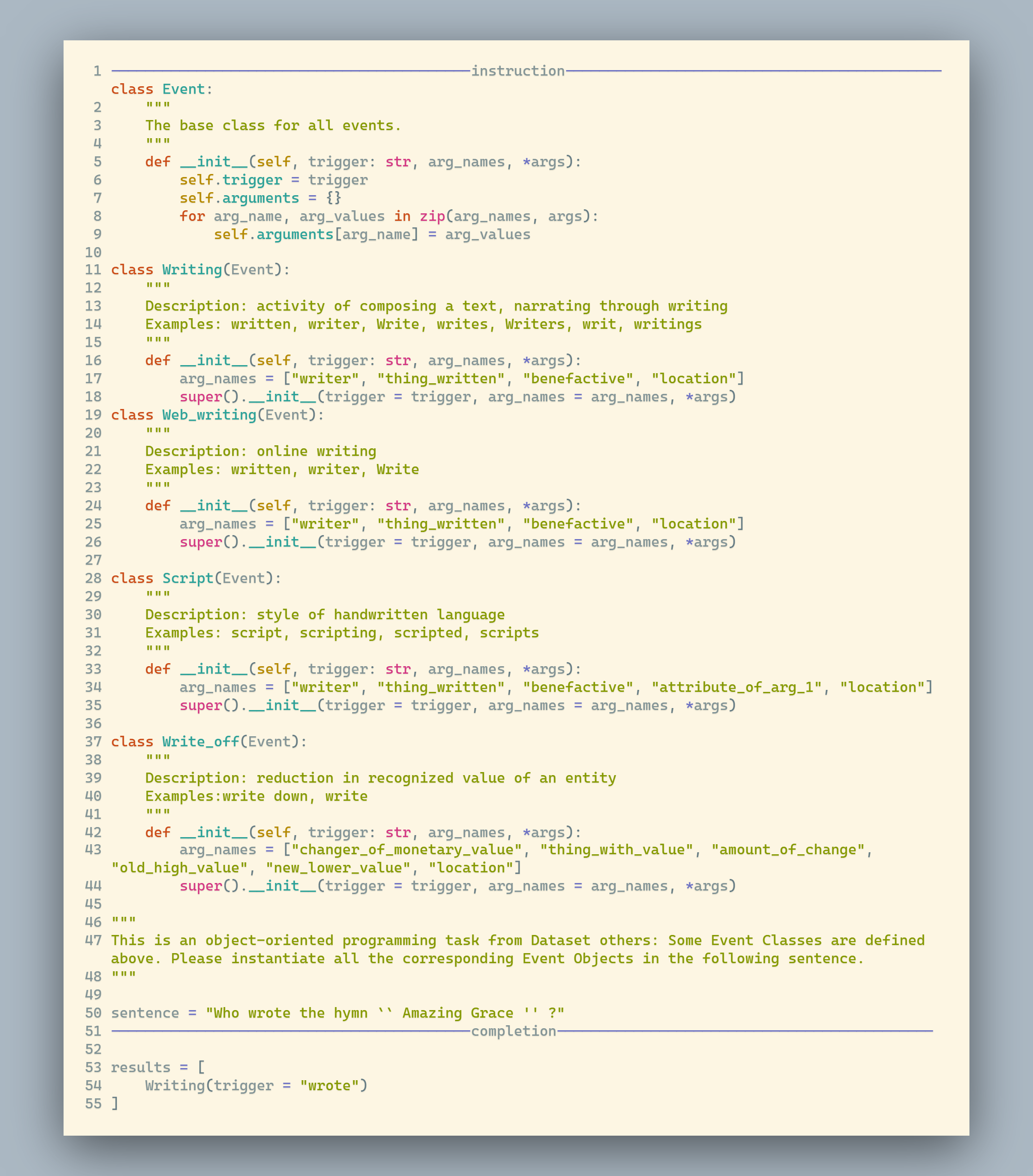}
   \caption{An example of training data in ED task.}
   \label{fig:ED_prompt}
\end{figure*}

\begin{figure*}[t]
   \centering
   \includegraphics[width=.9999\linewidth]{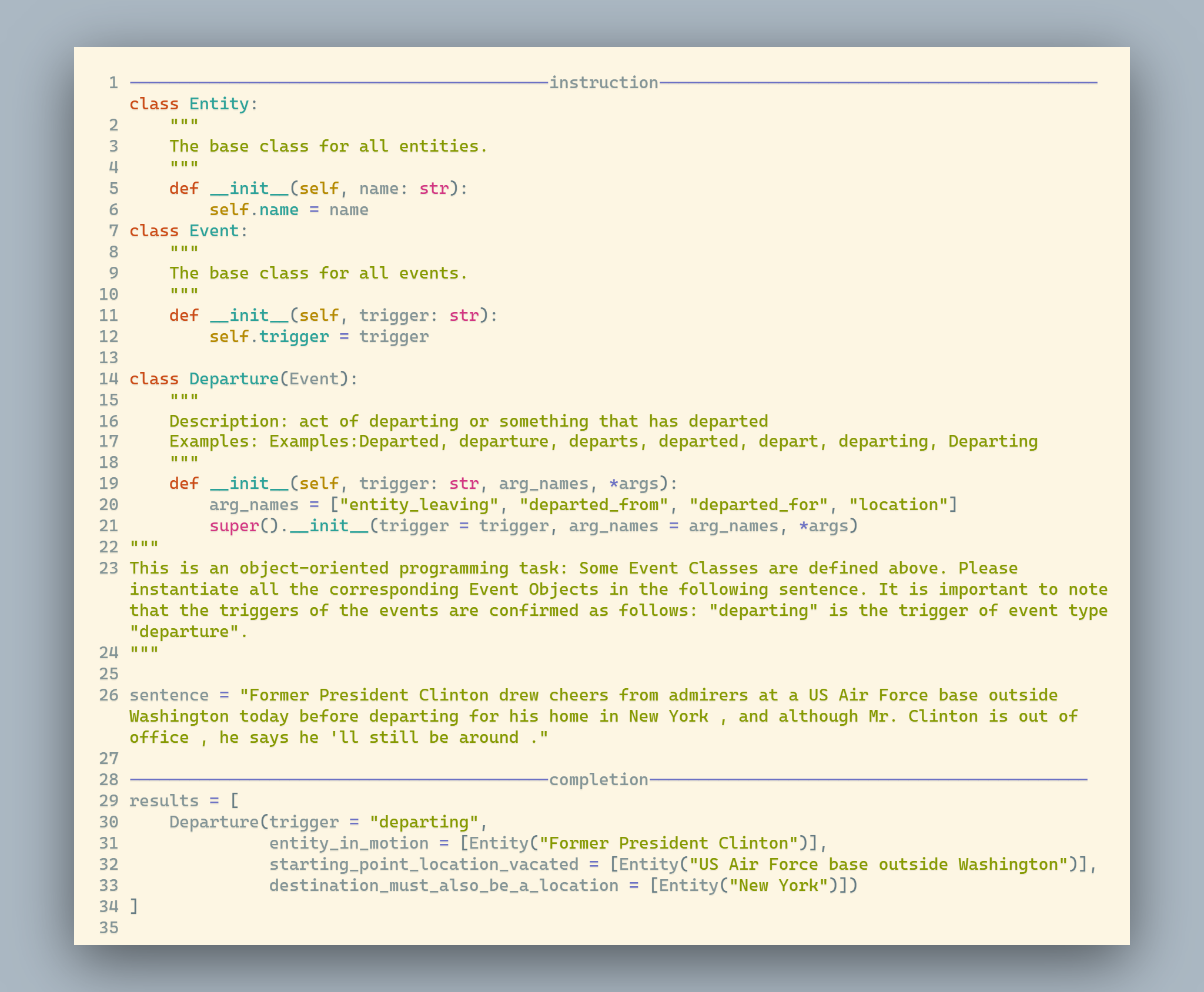}
   \caption{An example of training data in EAE task.}
   \label{fig:EAE_prompt}
\end{figure*}

We present examples of training prompts for both EAE and ED task, including instruction and completion in figure~\ref{fig:ED_prompt} and ~\ref{fig:EAE_prompt}.

More specifically, we employ the same Python class style prompt with KnowCoder~\cite{li2024knowcoder}, which includes the schema and instruction. Besides, we use the same output representation as the object of the certain class of event type.
Following the KnowCoder, we adopt class comments to provide clear definitions of concepts, which includes the event definition and several samples from the training set. 
However, while the KnowCoder use different schema representation for ED and EAE tasks, we use the same representation, which could better facilitate the model understanding about the certain event type.

\newpage
\begingroup
\begin{table*}[tp]
    \centering
    \begin{tabular}{p{\textwidth}}
        \toprule
        \underline{\textbf{\textsc{Prompt for Event Trigger Filtering.}}} \\
        \vspace{-2mm}
        \# Identify Reasonable Propbank Event Annotations in a Sentence
        \\ \\
        \#\# Task Overview:
        Your goal is to determine which **Propbank event annotations** are reasonable for the given sentence. If an annotation aligns with the context and semantics of the sentence, it is considered reasonable. Output the reasonable annotations in the specified format. If no reasonable annotations are found, output ``None of them.''
        \\ \\
        \#\# Guidelines: \\
1. **Understanding Event Annotations**:  
   Each Propbank event annotation consists of two parts: the **event trigger ** (outside the parentheses) and the **Propbank event type** (inside the parentheses). The trigger word is the specific word in the sentence that activates the event, while the Propbank event type defines the semantic category of the event. For example, in `robbery(rob.01)`, "robbery" is the trigger, and `rob.01` is the Propbank event type. Ensure you understand both components when evaluating the reasonableness of the annotation.

2. **Event Annotation Analysis**:  
   Review the definitions and usage patterns of the Propbank event annotations provided. Understand their semantic boundaries and structural characteristics.

3. **Sentence Context Evaluation**:  
   Assess whether the sentence provides sufficient context to support each Propbank event annotation. Consider whether the verb or action described in the sentence matches the annotation's definition.

4. **Reasonableness Assessment**:  
   Check if the Propbank event annotation is used appropriately in the given context and whether it matches the event definition in a clear and complete manner.
   
5. **Adverb/Adjective Trigger Exclusion**:  
   Be aware that some adverbs, adjectives, or partial verbs may be annotated as triggers by Propbank, leading to the generation of events. However, these annotations may not align with how humans typically define events. Such cases should be considered **unreasonable** and excluded from the final output.
\\ \\
\#\# Output Format:

Adhere to the specified format for clarity and consistency. Follow the **Example Output Format** to provide the **Reasonable Annotations** and **Reasoning**.
\\ \\
\#\# Example

\#\#\# Input:
**Sentence**: Generosity is voluntarily giving money on your own free will; robbery is taking money and giving it to others against their will.  \\
**Propbank Annotations**: `voluntarily(volunteer.01)`, `robbery(rob.01)`

\#\#\# Output: \\
**Reasonable Annotations**: `robbery(rob.01)`  \\
**Reasoning**:  \\
The sentence describes two contrasting concepts: generosity and robbery. The Propbank annotation `robbery(rob.01)` accurately captures the action of "taking money and giving it to others against their will," which aligns with the definition of robbery as an event involving theft or forced redistribution of property.  \\

However, `voluntarily(volunteer.01)` is not reasonable because "voluntarily" is an adverb describing the manner in which the action of giving occurs. It does not represent an event or action itself but rather modifies the act of generosity. Therefore, only `robbery(rob.01)` is a reasonable annotation.\\ \\

\#\# Input \\
**Sentence**: Five years ago they were negotiating with Milosevic at Dayton to stop the war.\\
**Propbank Annotations**: `negotiating(negotiate-01)`, `war(war-01)`

    \#\# Output \\

        \bottomrule
    \end{tabular}
    
    \caption{Prompt for Event Trigger Filtering.}
    \label{tab:trigger-filtering}
\end{table*}
\endgroup
\begingroup
\begin{table*}[tp]
    \centering

    \begin{tabular}{p{\textwidth}}
        \toprule
        \underline{\textbf{\textsc{Prompt for Event Type Refinement.}}} \\
\# Identify the Event Type Based on the Trigger Word \\ \\

\#\# Task Overview:
Your goal is to determine whether the given trigger matches any of the listed event types. If a match is found, output the serial number of the corresponding event type. If no match is found, output the last option.  \\ \\

\#\# Guidelines:

1. **Event Type Analysis:** Thoroughly review the definitions of all listed event types to grasp their semantic boundaries and structural characteristics.

2. **Trigger Evaluation:** Assess whether the trigger word aligns with the semantic and contextual requirements of any event type. If the trigger does not align with any event type, output the last option.

3. **Contextual Validation:** Analyze the sentence in which the trigger word appears to ensure the match is precise and meaningful.

4. **Context Reasonableness Analysis:** Check if the trigger word is used appropriately in the given sentence and whether it matches the event definition in a clear and complete manner.

5. **Annotation Accuracy Evaluation:** Assess whether the annotation of the trigger word is reasonable based on the event definition and sentence.

6. **Output Format:** Adhere to the specified format for clarity and consistency. Follow the Example Output Format to provide the Event Type and Reasoning.
\\ \\
\#\# Example

\#\#\# Input:
**Event Types**:
A. state\_crime: crime perpetrated by a sovereign state \\
B. war\_crime: serious violation of the laws of war\\
C. crime: unlawful act forbidden and punishable by criminal law\\
D. corporate\_crime: crimes committed either by a corporation or its representatives\\
E. environmental\_crime: illegal act which directly harms the environment\\
F. ethnic\_cleansing: systematic removal of a certain ethnic or religious group\\
G. alcohol-related\_crime: criminal activities that involve alcohol use\\
H. cybercrime: any crime that involves a computer and a network\\
I. criminal\_case: investigation case under criminal law\\
J. None of them.\\

**Trigger**: crime\\
**Sentence**: this is not because their race or culture is wrong or bad in any way, it 's simply a fact that certain groups have higher rates of crime.
\\  \\
\#\#\# Output: 
**Event Type**: C \\
**Reasoning**: The trigger word "crime" aligns with the core semantics of the crime event type (Event Type C). The context discusses crime in general terms without specifying a sovereign state (state\_crime), war-related violations (war\_crime), or any specific subtype such as corporate, environmental, or cybercrime. Thus, it best fits the generic "crime" definition. \\ \\

\#\# Input
**Event Types**: \\
A. negotiation: dialogue between two or more people or parties intended to reach a beneficial outcome \\
B. parley: type of diplomatic meeting held between enemies \\
C. None of them. \\

**Trigger**: crime \\
**Sentence**: Five years ago they were negotiating with Milosevic at Dayton to stop the war.

\#\# Output  \\
        
        \bottomrule
    \end{tabular}
    
    \caption{Prompt for Event Type Refinement.}
    \label{tab:type-refinement}
\end{table*}
\endgroup
\begingroup
\begin{table*}[tp]
    \centering
    \begin{tabular}{p{\textwidth}}
        \toprule
        \underline{\textbf{\textsc{Prompt for Event Argument Annotation.}}} \\
        \vspace{-2mm}
\# Annotating the Event Arguments based on the Schema.
\\ \\
\#\# Task Overview:
\\ \\
Your goal is to identify and annotation the arguments related to a specific event type in the given text. Each event type has specific roles (arguments), and you are required to annotation the arguments according to the provided event type and its corresponding argument roles. Do not introduce new roles or categories that are not part of the given event type and argument roles.

\#\# Guidelines:
\\ \\
1. **Understand Event Type and Argument Roles:** Before starting the annotation process, make sure you understand the provided event type and the roles (arguments) associated with it. Do not assume new roles or make connections outside the given structure. Only use the roles provided in the event type definition.

2. **Exact Match with Roles:** Each argument should align directly with one of the provided roles. For example:
   - If the event type is **Attack Event**, the roles are **Agent**, **Target**, **Location**, **Time**, **Cause**, and **Method**. You must only annotate arguments that fall under these categories. Do not annotate anything as **Purpose** or any role not explicitly listed.
   
3. **Contextual Accuracy:** Ensure that the arguments labeled in the text match the defined roles and are appropriate for the context. The roles should be matched based on both the meaning and the context of the text.

4. **Multiple Entities per Role:** Some roles may correspond to multiple entities. For example, **Agent** could involve more than one person or entity (e.g., `['A', 'B']`). In such cases, include all relevant entities under the role.

5. **Output Format:** Provide the annotated arguments in the following JSON format. Ensure that the roles and arguments strictly correspond to those defined for the given event type.
\\ \\
\#\# Example

\#\#\# Example Input:

**Event Type:** creation \\
**Event Description:** process during which something comes into being and gains its characteristics\\
**Trigger:** create\\
**Argument Roles:** agent\_creator, result\_thing\_created, material\_materials\_used, location\\
**Sentence:** they create a market for themselves .\\

\#\#\# Example Output:

```json
\{\\
    "agent\_creator": ["drug pushers"],\\
    "result\_thing\_created": ["market"],\\
    "material\_materials\_used": [],\\
    "location": [],\\
\}\\
```json
\\ \\
\#\# Input \\
**Event Type:** parley \\
**Event Description:** type of diplomatic meeting held between enemies \\
**Trigger:** negotiating\\
**Argument Roles:** Negotiator, Other party, Location\\
**Sentence:** Five years ago they were negotiating with Milosevic at Dayton to stop the war. \\

        \bottomrule
    \end{tabular}
    
    \caption{Prompt for Event Argument Annotation.}
    \label{tab:argument-annotation}
\end{table*}
\endgroup
\begingroup
\begin{table*}[tp]
    \centering
    \begin{tabular}{p{\textwidth}}
        \toprule
        \underline{\textbf{\textsc{Prompt for Event Offset Alignment.}}} \\
        \vspace{-2mm}
\# Refining the Annotation with the Rules \\ \\

\#\# Task Overview:
You will be provided with an initial event argument extraction result. Your task is to analyze this result according to the given annotation rules. Based on the rules, you will need to ensure the arguments are correctly selected or re-generate the correct extraction if necessary. The final output should be a **single JSON** string containing the event arguments, formatted according to the rules. \\ \\

\#\# Rules for Evaluation and Modification: 

1. **Logical Consistency:** 
   - Ensure that each argument logically participates in the event. For example, the **Agent** typically refers to the entity executing the trigger, while the **Entity** refers to the recipient or affected party of the trigger.

2. **Accuracy of Extraction:**
   - Arguments must be extracted concisely and clearly. If a role is extracted as "A and B," it should be represented as `["A", "B"]` rather than a combined form.
   - Avoid redundancy or unnecessary details.
   - The arguments must be extracted from the context.

3. **Conciseness:**
   - Arguments should be represented as single words or phrases. If the extraction includes a sentence, reduce it to the smallest meaningful segment (e.g., remove unnecessary words like "that").

4. **Avoid Redundant or Overlapping Arguments:**
   - Check for any overlapping or redundant arguments. If an overlap exists, determine whether it is logical or whether one of the arguments should be removed or merged.

5. **Argument Selection:**
   - Use the provided **Input** to determine which argument is the most accurate or appropriate for each role.
   - In cases where the input is not fully accurate, modify or re-generate the argument extraction to match the correct role.

6. **Final Output Format:**
   - Provide the final output in **JSON format**, with each event role and its arguments clearly defined. Do not include the event type or trigger word.
\\ \\
\#\# Example:

\#\#\# Example Input:

**Event Type:**\\
- **Event Type:** Attack\\
- **Event Definition:** action to injure another organism

**Event Trigger:**
attacked

**Event Roles Definition:**\\
- **Roles:** Agent, Location, Time, Target

**Sentence:**
John and Sarah attacked the enemy base at night.

**Input:** \\
\{
  "Agent": ["John and Sarah"],
  "Target": ["the enemy base"],
  "Time": ["at night"],
  "Location": ["the enemy base"]
\}

\#\#\# Example Output:\\
\{
  "Agent": ["John", "Sarah"],
  "Target": ["enemy base"],
  "Time": ["night"],
  "Location": []
\}

\#\# Input \\
**Event Type:** \\
- **Event Type:** parley \\
- **Event Definition:** type of diplomatic meeting held between enemies

**Event Trigger:**
negotiating

**Event Roles Definition:** \\
- **Roles:** Negotiator, Other party, Location

**Sentence:**
Five years ago they were negotiating with Milosevic at Dayton to stop the war.

**Input:**

\{
  "Negotiator": ["They"],
  "Other party": ["Milosevic"],
  "Location": ["at Dayton"],
\} \\
\#\# Output \\

        \bottomrule
    \end{tabular}
    
    \caption{Prompt for Event Offset Alignment.}
    \label{tab:argument-preference}
\end{table*}
\endgroup
\begingroup
\begin{table*}[tp]
    \centering
    \begin{tabular}{p{\textwidth}}
        \toprule
        \underline{\textbf{\textsc{Prompt for Event Offset Alignment with multiple input.}}} \\
        \vspace{-2mm}
\#\# Task Overview:
You will be provided with three initial event argument extraction results, named **Input1**, **Input2**, and **Input3**. Your task is to analyze and combine these results according to the given annotation rules. Based on the rules, you will need to choose the most appropriate argument for each role or re-generate the correct extraction if necessary. The final output should be a **single JSON** string containing the event arguments, formatted according to the rules.

\#\# Rules for Evaluation and Modification:

1. **Logical Consistency:** 
   - Ensure that each argument logically participates in the event. For example, the **Agent** typically refers to the entity executing the trigger, while the **Entity** refers to the recipient or affected party of the trigger.

2. **Accuracy of Extraction:**
   - Arguments must be extracted concisely and clearly. If a role is extracted as "A and B," it should be represented as `["A", "B"]` rather than a combined form.
   - Avoid redundancy or unnecessary details.

3. **Conciseness:**
   - Arguments should be represented as single words or phrases. If the extraction includes a sentence, reduce it to the smallest meaningful segment (e.g., remove unnecessary words like "that").

4. **Avoid Redundant or Overlapping Arguments:**
   - Check for any overlapping or redundant arguments. If an overlap exists, determine whether it is logical or whether one of the arguments should be removed or merged.

5. **Argument Selection:**
   - Compare **Input1**, **Input2**, and **Input3** to determine which argument is the most accurate or appropriate for each role. If there is disagreement, use the rules to guide which argument should be selected.
   - In cases where none of the inputs is fully accurate, combine or re-generate the most accurate argument extraction.

6. **Final Output Format:**
   - Provide the final output in **JSON format**, with each event role and its arguments clearly defined. Do not include the event type or trigger word.

---

\#\#\# Input Format:
You will be given:
- **Input1, Input2, Input3:** Three JSON objects containing the initial event argument extractions.
- **Event Roles Definition:** A list of roles defined for the event (e.g., Agent, Target, Location, etc.).
- **Context:** The original text from which the arguments are extracted.

\#\#\# Output Format:
A **single JSON object** with the final, modified extraction, including the most accurate and logically consistent arguments for each role.

\#\#\# Example:

\#\#\#\# Input:

**Event Roles Definition:**
- **Roles:** Agent, Location, Time, Target

**Context:**
"John and Sarah attacked the enemy base at night."

**Input1:**
```json
{
  "Agent": ["John"],
  "Location": ["enemy base"],
  "Time": ["at night"],
  "Target": []
}
'''
**Input2:**
```json
{
  "Agent": ["John", "Sarah"],
  "Location": ["enemy base"],
  "Time": ["night"],
}
'''

**Input3:**
```json
{
  "Agent": ["John and Sarah"],
  "Location": ["enemy base"],
  "Time": ["night"],
}
'''
**Output:**
```json
{
  "Agent": ["John", "Sarah"],
  "Location": ["enemy base"],
  "Time": ["night"],
  "Target": []
}
'''

**Explanation:**

-The Agent role was chosen as ["John", "Sarah"] since both individuals participated in the action.

-The Location and Time were correctly extracted as "enemy base" and "night," respectively.
-The Target role remains empty as no direct target is mentioned in the context. \\

        \bottomrule
    \end{tabular}
    
    \caption{Prompt for Event Offset Alignment with multiple input.}
    \label{tab:argument-annotation-multi}
\end{table*}
\endgroup

\newpage


\end{document}